\documentclass[11pt,letterpaper]{article}
\usepackage{naaclhlt2016}
\usepackage{times}
\usepackage{latexsym}
\usepackage{xcolor}
\usepackage{graphicx}

\definecolor{nice-red}{HTML}{E41A1C}
\definecolor{nice-orange}{HTML}{FF7F00}
\definecolor{nice-yellow}{HTML}{FFC020}
\definecolor{nice-green}{HTML}{4DAF4A}
\definecolor{nice-blue}{HTML}{377EB8}
\definecolor{nice-purple}{HTML}{984EA3}

\naaclfinalcopy 
\def\naaclpaperid{***} 

\usepackage{amssymb}
\usepackage{amsmath}
\usepackage{bm}

\newcommand{\Joh}[1]{{\color{nice-green}JW: #1}}

\newcommand{\R}{\mathbb{R}}

\title{A Factorization Machine Framework for Testing Bigram Embeddings in Knowledgebase Completion}

\author{Johannes Welbl{\normalfont,}~ Guillaume Bouchard \and Sebastian Riedel\\
         University College London\\
         London, UK\\
         \small{\tt \{j.welbl, g.bouchard, s.riedel\}@cs.ucl.ac.uk}}

\date{}

\begin{document}

\maketitle

\begin{abstract}
Embedding-based Knowledge Base Completion models have so far mostly combined distributed representations of \emph{individual} entities or relations to compute truth scores of missing links. Facts can however also be represented using pairwise embeddings, i.e. embeddings for \emph{pairs} of entities and relations. In this paper we explore such bigram embeddings with a flexible Factorization Machine model and several ablations from it. We investigate the relevance of various 
bigram types on the \texttt{fb15k237} dataset 
and find relative improvements compared to a compositional model. 
\end{abstract}

\section{Introduction}
    Present day Knowledge Bases (KBs) such as 
    YAGO~\cite{Suchanek2007:YAGO}, Freebase~\cite{Bollacker2008:FB} or the Google Knowledge Vault~\cite{Dong:2014:KnowledgeVault} provide immense collections of structured knowledge. Relationships in these KBs often exhibit regularities and models that capture these can be used to predict missing KB entries.
    A common approach to KB completion is via tensor factorization, where a collection of fact triplets is represented as a sparse mode-$3$ tensor which is decomposed into several low-rank sub-components.
    Textual relations, i.e. relations between entity pairs extracted from text, can aid the imputation of missing KB facts by modelling them together with the KB relations~\cite{Riedel2013:Universal}.

    The general merit of factorization methods for KB completion has been demonstrated by a variety of models, such as RESCAL~\cite{Nickel2011:RESCAL}, TransE~\cite{Bordes2013:TransE} and DistMult~\cite{Yang2014:DistMult}.
    These models learn distributed representations for entities and relations (be it as vector or as matrix) and infer the truth value of a fact by combining embeddings for these constituents in an appropriate composition function.

    Most of these factorization models however operate on the level of embeddings for single entities and relations. The implicit assumption here is that facts are \emph{compositional}, i.e. that the subject, relation and object of a fact are its atomic constituents. Semantic aspects relevant for imputing its truth can directly be recovered from its constituents when composing their respective embeddings in a score.

    For further notation let $\texttt{E}$ and $\texttt{R}$ be sets of entities and relations, respectively. We denote a fact $f$ stating a relation $r \in \texttt{R}$ between subject $s \in \texttt{E}$ and object $o \in \texttt{E}$ as $f=(s,r,o)$.
    Our goal is to learn embeddings for larger sub-constituents of $f$ than just $s,r$, and $o$: we want to learn embeddings also for the entity pair bigram $(s,o)$ as well as the relation-entity bigrams 
    $(s,r)$ and $(r,o)$. 
    As an example, consider Freebase facts with relation \texttt{eating/practicer\_of\_diet/diet} and object \texttt{Veganism}. Overall only two objects are observed for this relation and it thus makes sense to learn a joint embedding for bigrams 
    $(r,o)$ together, instead of distinct embeddings for each atom alone and then having to learn their compatibility.

    While \newcite{Riedel2013:Universal} have trained embeddings only for entity pairs, we will in this paper explore the role of general bigram embeddings for KB completion, i.e. also the embeddings for other possible pairs of entities and relations. This is achieved using a Factorization Machine (FM) framework~\cite{Rendle2010:FM} that is modular in its feature components, allowing us to selectively add or discard certain bigram embeddings and compare their relative importance. All models are empirically compared and evaluated on the \texttt{fb15k237} dataset from~\newcite{Tout2015}. 

    In summary, our main contributions are: 
    i)~Adressing the question of generic bigram embeddings in a KB completion model for the first time; ii)~The adaption of Factorization Machines for this matter; iii) Experimental findings for comparing  different bigram embedding models on \texttt{fb15k237}.  

\section{Related Work}

    In the Universal Schema model (model \emph{F}),~\newcite{Riedel2013:Universal} factorize KB entries together with relations of entity pairs extracted from text, embedding textual relations in the same vector space as KB relations. 
    \newcite{Singh2015:Combined} extend this model to include a variety of other interactions between entities and relations, using different relation vectors to interact with subject, object or both.
    \newcite{Jennatton2012:Multi} also recognize the need to integrate rich higher-order interaction information into the score. Like~\newcite{Nickel2011:RESCAL} however, their model specifies relationships as relation-specific bilinear forms of entity embeddings. Other embedding methods for KB completion include DistMult~\cite{Yang2014:DistMult} with a trilinear score, and TransE~\cite{Bordes2013:TransE} which offers an intriguing geometrical intuition.
    Among the aforementioned methods, embeddings are mostly learned for individual subjects, relations or objects; merely model $F$ ~\cite{Riedel2013:Universal} constitutes the exception. 

    Some methods rely on more expressive composition functions to deal with non-compositionality or interaction effects, such as the  Neural Tensor Networks~\cite{Socher2013:NTN} or the recently introduced Holographic Embeddings~\cite{Nickel2015:HolE}. In comparison to the otherwise used (generalized) dot products, the composition functions of these models enable richer interactions between unit constituent embeddings. However, this comes with the potential disadvantage of presenting less well-behaved optimisation problems and being slower to train.
    Factorization Machines have already been applied in a similar setting to ours by~\newcite{Petroni2015:CORE} who use them with contextual features for an Open Relation Extraction task, but without bigrams.



\section{Model}
\subsection{Brief Recall of Factorization Machines}
A Factorization Machine (FM) is a quadratic regression model with low-rank constraint on the quadratic interaction terms.\footnote{We disregard the more general extension to higher-order interactions that is described in the original FM paper  and only consider the quadratic case. Also, we omit the global model bias as we found that it was not helpful for our task empirically.} 
Given a sparse input feature vector $\bm{\phi}=(\phi_1,\dots,\phi_n)^T \in \R^{n}$, the FM output prediction $X \in \R$ is
\begin{equation}\label{eq:FM_basic}
    X = \left<\mathbf{v}, \bm{\phi}\right> + \sum_{i,j=1}^{n} \left<\mathbf{w}_i,\mathbf{w}_j\right> \cdot \phi_i \phi_j
\end{equation}
 where $\mathbf{v} \in \R^{n}$, and $\forall i,j=1,\dots n:~ \mathbf{w}_i,\mathbf{w}_j\in \R^k $ are model parameters with $k \ll n$ and $\left<\cdot,\cdot\right>$ denotes the dot product.
 Instead of allowing for an individual quadratic interaction coefficient per pair $(i,j)$, the FM assumes that the matrix of quadratic interaction coefficients has low rank $k$; thus the interaction coefficient for feature pair $(i,j)$ is represented by an inner product of $k$-dimensional vectors $\mathbf{w}_i$ and $\mathbf{w}_j$. The low rank constraint (i) provides a strong form of regularisation to this otherwise over-parameterized model, (ii) pools statistical strength for estimating similarly profiled interaction coefficients and (iii) retains a total number of parameters linear in $n$.
In summary, with a FM one can efficiently harness a large set of sparse features and interactions between them while retaining linear memory complexity.
\subsection{Feature Representation for Facts}
For the KB completion task we will use a FM with unit and bigram indicator features to learn low-rank embeddings for both.
To formalize this, we will refer to the elements of the set $\mathtt{U}_f = \{s,r,o\}$ as \emph{units} of fact $f$, and to the elements of $\mathtt{B}_f = \{(s,r), (r,o), (o,s)\}$ as 
\emph{bigrams}
of fact $f$. Let $\iota_u \in \R^{|\texttt{E}|+|\texttt{R}|}$ be the one-hot indicator vector that encodes a particular unit\footnote{For subject and object the same entity embedding is used.} $u \in (\texttt{E} \cup \texttt{R})$.
Furthermore we define $\iota_{(s,r)}\in \R^{|\texttt{E}||\texttt{R}|}$,  $\iota_{(r,o)} \in \R^{|\texttt{R}||\texttt{E}|}$ and  $\iota_{(o,s)} \in \R^{|\texttt{E}|^2}$ to be the one-hot indicator vectors encoding particular bigrams.
Our feature vector $\bm{\phi}(f)$ for fact $f=(s,r,o)$ then consists of simply the concatenation of indicator vectors for all its units and bigrams:
\begin{equation}\label{features}
    \bm{\phi}(f) = \text{concat}(\iota_s, \iota_r, \iota_o, \iota_{(s,r)}, \iota_{(r,o)}, \iota_{(o,s)})
\end{equation}
This sparse set of features provides a rich representation of a fact with indicators for subject, relation and object, as well as any pair thereof. 

\subsection{Scoring a Fact}
Harnessing the expressive benefits of a sigmoid link function for relation modelling~\cite{bouchard2015:logistic}, we define the truth score of a fact as 
$g(f) = \sigma(X_{f})$
where $\sigma$ is the sigmoid function and $X_{f}$ is given as output of the FM model (\ref{eq:FM_basic}) with unit and bigram features $\bm{\phi}(f)$ as defined in (\ref{features}): 
\begin{equation}
\label{eq:score}
    X_{f} = \left< \bm{\phi}(f), \mathbf{v} \right> +  \sum_{  i,j=1 }^{n} \left<\mathbf{w}_{i}, \mathbf{w}_{j}\right> \cdot \phi_i(f) \phi_j(f)
\end{equation}
Since our feature vector $\bm{\phi}(f)$ is sparse with only six active entries, we can re-express (\ref{eq:score}) in terms of the activated embeddings which we directly index by their respective units and bigrams:
\begin{equation}
\label{eq:score_sparse}
    X_{f} = \sum_{c \in (\mathtt{U}_f \cup \mathtt{B}_f)}v_c +  \sum_{  c_1, c_2  \in (\mathtt{U}_f \cup \mathtt{B}_f) } \left<\mathbf{w}_{c_1}, \mathbf{w}_{c_2}\right>
\end{equation}
This score comprises all possible interactions between any of the units and bigrams of $f$.

\subsection{Model Ablations for Investigating Particular Bigram Embeddings}
The score (\ref{eq:score_sparse}) can easily be modified and individual summands removed from it. In particular, when discarding all but one summand, model \emph{F} is recovered, i.e. with $c_1=(s,o)$; $c_2=r$. 
On the other hand, alternatives to model \emph{F} with other bigrams than entity pairs can be tested
by removing all summands but the one of a single bigram $b \in B_f$~ vs. the remaining complementary unit $u \in U_f$:
\begin{equation}
\label{eq:score_altered2}
    X_{f}^{u,b} = \mathbf{v}_u + \mathbf{v}_b +  \left<\mathbf{w}_{u}, \mathbf{w}_{b}\right>
\end{equation}
This general formulation offers us a method for investigating the relative impact of all combinations of bigram vs. unit embeddings besides model $F$, namely the models with $u = s$; $~b=(r,o)$ and with $u=o$; $~b=(s,r)$.

\subsection{Training Objective}
Given sets of true training facts $\Omega^+$ and sampled negative facts $\Omega^-$, we minimize the following loss:
\begin{equation}
  -\sum_{f \in \Omega^+}^{} \log(1+e^{X_f}) + \frac{1}{\eta}\sum_{f\in\Omega^-}^{} \log(1+e^{X_f})
    \label{eq:training-objective}
\end{equation}
where the parametrization of $X_f$ is learned.
 We use the hyperparameter $\eta\in\R^+$ for denoting the ratio of negative facts that are sampled per positive fact so that the contributions of true and false facts are balanced even if there are more negative facts than positives. The loss differs from a standard negative log-likelihood objective with  logistic link, but we found that it performs better in practice. The intuition comes from the fact that instead of penalizing badly classified positive facts, we put more emphasis (i.e. negative loss) on positive facts that are correctly classified. Since we used an $L_2$ regularization 
 and the loss is asymptotically linear, the resulting objective is continuous and bounded from below, guaranteeing a well defined local minimum. 

\section{Experiments}
The bigram embedding models are tested on \texttt{fb15k237}~\cite{Tout2015}, a dataset comprising both Freebase facts and lexicalized dependency path relationships between entities.


\begin{table*}[t]
    \centering
    \footnotesize
    \begin{tabular}{@{\extracolsep{0pt}}r   l    r  rrr | rrr@{}}
        &&& \multicolumn{3}{c}{overall \emph{HITS@}} & \multicolumn{3}{c}{\emph{MRR}} \\
        \cline{4-6} \cline{7-9}

        &\textbf{Model} & $\tau$ &                    1 & 3 & 10 & overall & no TM & with TM\\
        \cline{2-9}
        &DistMult & 0.0         & 18.2 & 27.0 & 37.9 & 24.8 & 28.0 & 16.2  \\
        &full FM & 0.0 &   20.1 & 28.7 & 38.9 & 26.4 & 29.3 & 18.3  \\
        (*)& $(s,o)$ vs. $r$ & 1.0           & 2.1 & 3.8 & 6.5 & 3.5 & 0.0 & 13.1  \\
        (**)& $(r,o)$ vs. $s$  &    0.1     & 24.9 & 34.8 & 45.8 & 32.0 & 34.7 & 24.8  \\
        (***)&$(s,r)$ vs. $o$  &     0.0 & 9.0   & 17.3 & 29.9 & 15.6 & 17.3 & 10.9   \\
        &(*)~+~(**)~+~(***) & 0.1 & \bf{25.9} & \bf{36.2} & \bf{47.4} & \bf{33.2} & \bf{35.0} & \bf{28.3} \\
    \end{tabular}
    \caption{Test set metrics for different models and varying unit and bigram embeddings on \texttt{fb15k237}, all performance numbers in \% and best result in bold. The optimal value for $\tau$ is indicated as well. 
    }
    \label{results}
\end{table*}

\paragraph{Training Details and Evaluation}
We optimized the loss using AdaM~\cite{Kingma2014:Adam} with minibatches of size 1024, using initial learning rate 1.0 and initialize model parameters from $\mathcal{N}(0,1)$. Furthermore, a hyperparameter $\tau <1 $ like in~\cite{Tout2015} is introduced to discount the importance of textual mentions in the loss.
When sampling a negative fact we alter the object of a given training fact $(s,r,o)$ at random to $o' \in \mathtt{E}$, and repeat this $\eta$ times, sampling negative facts every epoch anew. There is a small implied risk of sampling positive facts as negative, but this is rare and the discounted loss weight of negative samples mitigates the issue further.
Hyperparameters ($L_2$-regularisation, $\eta$, $\tau$, latent dimension $k$) are selected in a grid search for minimising Mean Reciprocal Rank (\emph{MRR}) on a fixed random subsample of size 1000 of the validation set. All reported results are for the test set. 
We use the competitive unit model DistMult as baseline and employ the same ranking evaluation scheme as in~\cite{Tout2015} and~\cite{toutanova2015_observed}, computing filtered \emph{MRR} and \emph{HITS} scores whilst ranking true test facts among candidate facts with altered object. Particular bigrams that have not been observed during training have no learned embedding; a 0-embedding is used for these. This nullifies their impact on the score and models the back-off to using nonzero embeddings. 

\paragraph{Results}
Table \ref{results} gives an overview of the general results for the different models. Clearly, some of the bigram models can obtain an improvement over the unit DistMult model. In a more fine-grained analysis of model performances, characterized by whether entity pairs of test facts had textual mentions available in training (\emph{with TM}) or not (\emph{without TM}), the results exhibit a similar pattern like in~\cite{Tout2015}: most models perform worse on test facts with TM, only model $F$, which can learn very little without relations has a reversed behavior. A side observation is that several models achieved highest overall MRR with $\tau=0$, i.e. when not using TM.

The sum of the three more light-weight bigram models performs better than the full FM, even though the same types of embeddings are used. A possible explanation is that applying the same embedding in several interactions with other embeddings (as in the full FM) instead of only one interaction (like in (*)+(**)+(***)) makes it harder to learn since its multiple functionalities are competing.

Another interesting finding is that some bigram types achieve much better results than others, in particular model (**). A possible explanation becomes apparent with closer inspection of the test set: a given test fact $f$ usually contains at least one bigram $b \in \texttt{B}_f$ which has never been observed yet. In these cases the bigram embedding is 0 by design and only the offset values are used.
The proportions of test facts for which this happens are $73$\%, $10$\% and $24$\% respectively for the bigrams $(s,o)$, $(r,o)$, and $(s,r)$. Thus models (**) and (***) already have a definite advantage over model (*) that originates purely from the nature of the data. A trivial but somehow important lesson we can learn from this is that if we know about the relative prevalence of different bigrams (or more generally: sub-tuples) in our dataset, we can incorporate and exploit this in the sub-tuples we choose. 

Finally, for the initial example with relation \texttt{eating/practicer\_of\_diet/diet} and object \texttt{Veganism}, we indeed find that in all instances model (**) with its $(r,o)$ embedding gives the correct fact in the top 2 predictions, while the purely compositional DistMult model ranks it far outside the top 10. More generally, cases in which only a single object co-appeared with a test fact relation during training had  $95,3$\% \emph{HITS@1} with model (**) while only $52,6$\% for DistMult. This supports the intuition that bigram embeddings of $(r,o)$ are in fact better suited for cases in which very few objects are possible for a relation.

\section{Conclusion}
We have demonstrated that FM provide an approach to KB completion that can incorporate embeddings for bigrams naturally. The FM offers a compact unified framework in which various tensor factorization models can be expressed, including model F.

Extensive experiments have demonstrated that bigram models can improve prediction performances substantially over more straightforward unigram models.
A surprising but important result is that bigrams other than entity pairs are particularly appealing.

The bigger question behind our work is about compositionality vs. non-compositionality in a broader class of knowledge bases involving higher order information such as time, origin or context in the tuples. Deciding which modes should be merged into a high order embedding without having to rely on heavy cross-validation is an open question.

\section*{Acknowledgments}
We thank Th\'eo Trouillon, Tim Rockt\"aschel, Pontus Stenetorp and Thomas Demeester for discussions and hints, as well as the reviewers for comments. This work was supported by an EPSRC studentship, an Allen Distinguished Investigator Award and a Marie Curie Career Integration Award.

\bibliography{naaclhlt2016}

\begin{thebibliography}{}

\bibitem[\protect\citename{Bollacker \bgroup et al.\egroup
  }2008]{Bollacker2008:FB}
Kurt Bollacker, Colin Evans, Praveen Paritosh, Tim Sturge, and Jamie Taylor.
\newblock 2008.
\newblock {Freebase: a collaboratively created graph database for structuring
  human knowledge}.
\newblock In {\em SIGMOD 08 Proceedings of the 2008 ACM SIGMOD international
  conference on Management of data}, pages 1247--1250.

\bibitem[\protect\citename{Bordes \bgroup et al.\egroup
  }2013]{Bordes2013:TransE}
Antoine Bordes, Nicolas Usunier, Alberto Garcia-Dur\'an, Jason Weston, and
  Oksana Yakhnenko.
\newblock 2013.
\newblock Translating embeddings for modeling multi-relational data.
\newblock In {\em NIPS 26}.

\bibitem[\protect\citename{Bouchard \bgroup et al.\egroup
  }2015]{bouchard2015:logistic}
Guillaume Bouchard, Sameer Singh, and Theo Trouillon.
\newblock 2015.
\newblock On approximate reasoning capabilities of low-rank vector spaces.
\newblock In {\em AAAI Spring Syposium on Knowledge Representation and
  Reasoning (KRR): Integrating Symbolic and Neural Approaches}.

\bibitem[\protect\citename{Dong \bgroup et al.\egroup
  }2014]{Dong:2014:KnowledgeVault}
Xin Dong, Evgeniy Gabrilovich, Geremy Heitz, Wilko Horn, Ni~Lao, Kevin Murphy,
  Thomas Strohmann, Shaohua Sun, and Wei Zhang.
\newblock 2014.
\newblock Knowledge vault: A web-scale approach to probabilistic knowledge
  fusion.
\newblock In {\em Proceedings of the 20th ACM SIGKDD International Conference
  on Knowledge Discovery and Data Mining}, KDD '14, pages 601--610, New York,
  NY, USA. ACM.

\bibitem[\protect\citename{Jenatton \bgroup et al.\egroup
  }2012]{Jennatton2012:Multi}
Rodolphe Jenatton, Nicolas~L. Roux, Antoine Bordes, and Guillaume~R Obozinski.
\newblock 2012.
\newblock A latent factor model for highly multi-relational data.
\newblock In {\em NIPS 25}, pages 3167--3175. Curran Associates, Inc.

\bibitem[\protect\citename{Kingma and Ba}2015]{Kingma2014:Adam}
Diederik~P. Kingma and Jimmy Ba.
\newblock 2015.
\newblock Adam: {A} method for stochastic optimization.
\newblock {\em The International Conference on Learning Representations
  (ICLR)}.

\bibitem[\protect\citename{Nickel \bgroup et al.\egroup
  }2011]{Nickel2011:RESCAL}
Maximilian Nickel, Volker Tresp, and Hans peter Kriegel.
\newblock 2011.
\newblock A three-way model for collective learning on multi-relational data.
\newblock In Lise Getoor and Tobias Scheffer, editors, {\em Proceedings of the
  28th International Conference on Machine Learning (ICML-11)}, pages 809--816,
  New York, NY, USA. ACM.

\bibitem[\protect\citename{Nickel \bgroup et al.\egroup }2015]{Nickel2015:HolE}
Maximilian Nickel, Lorenzo Rosasco, and Tomaso Poggio.
\newblock 2015.
\newblock {Holographic Embeddings of Knowledge Graphs}.
\newblock Technical report, arXiv, October.

\bibitem[\protect\citename{Petroni \bgroup et al.\egroup
  }2015]{Petroni2015:CORE}
Fabio Petroni, Luciano~Del Corro, and Rainer Gemulla.
\newblock 2015.
\newblock Core: Context-aware open relation extraction with factorization
  machines.
\newblock In Lluís Màrquez, Chris Callison-Burch, Jian Su, Daniele Pighin,
  and Yuval Marton, editors, {\em EMNLP}, pages 1763--1773. The Association for
  Computational Linguistics.

\bibitem[\protect\citename{Rendle}2010]{Rendle2010:FM}
Steffen Rendle.
\newblock 2010.
\newblock Factorization machines.
\newblock In {\em Data Mining (ICDM), 2010 IEEE 10th International Conference
  on}, pages 995--1000. IEEE.

\bibitem[\protect\citename{Riedel \bgroup et al.\egroup
  }2013]{Riedel2013:Universal}
Sebastian Riedel, Limin Yao, Benjamin~M. Marlin, and Andrew McCallum.
\newblock 2013.
\newblock Relation extraction with matrix factorization and universal schemas.
\newblock In {\em Joint Human Language Technology Conference/Annual Meeting of
  the North American Chapter of the Association for Computational Linguistics
  (HLT-NAACL '13)}, June.

\bibitem[\protect\citename{Singh \bgroup et al.\egroup
  }2015]{Singh2015:Combined}
Sameer Singh, Tim Rocktaschel, and Sebastian Riedel.
\newblock 2015.
\newblock Towards combined matrix and tensor factorization for universal schema
  relation extraction.
\newblock In {\em NAACL Workshop on Vector Space Modeling for NLP}.

\bibitem[\protect\citename{Socher \bgroup et al.\egroup }2013]{Socher2013:NTN}
Richard Socher, Danqi Chen, Christopher~D Manning, and Andrew Ng.
\newblock 2013.
\newblock Reasoning with neural tensor networks for knowledge base completion.
\newblock In {\em NIPS 26}.

\bibitem[\protect\citename{Suchanek \bgroup et al.\egroup
  }2007]{Suchanek2007:YAGO}
Fabian~M. Suchanek, Gjergji Kasneci, and Gerhard Weikum.
\newblock 2007.
\newblock {YAGO:} a core of semantic knowledge unifying {WordNet} and
  {Wikipedia}.
\newblock In {\em WWW '07: Proceedings of the 16th International World Wide Web
  Conference, Banff, Canada}, pages 697--706.

\bibitem[\protect\citename{Toutanova and Chen}2015]{toutanova2015_observed}
Kristina Toutanova and Danqi Chen.
\newblock 2015.
\newblock Observed versus latent features for knowledge base and text
  inference.
\newblock In {\em Workshop on Continuous Vector Space Models and Their
  Compositionality (CVSC)}.

\bibitem[\protect\citename{Toutanova \bgroup et al.\egroup }2015]{Tout2015}
Kristina Toutanova, Danqi Chen, Patrick Pantel, Hoifung Poon, Pallavi
  Choudhury, and Michael Gamon.
\newblock 2015.
\newblock Representing text for joint embedding of text and knowledge bases.
\newblock In {\em Empirical Methods in Natural Language Processing (EMNLP)}.
  ACL – Association for Computational Linguistics, September.

\bibitem[\protect\citename{Yang \bgroup et al.\egroup }2014]{Yang2014:DistMult}
Bishan Yang, Wen{-}tau Yih, Xiaodong He, Jianfeng Gao, and Li~Deng.
\newblock 2014.
\newblock Embedding entities and relations for learning and inference in
  knowledge bases.
\newblock {\em CoRR}, abs/1412.6575.

\end{thebibliography}
\bibliographystyle{naaclhlt2016}

\end{document}